\documentclass[a4paper,11pt]{article}
\usepackage{jinstpub_edited}
\usepackage{lineno}

\usepackage{graphics}
\usepackage{graphicx}%
\usepackage{multirow}%
\usepackage{amsmath,amsfonts}%
\usepackage{amsthm}%
\usepackage{mathrsfs}%
\usepackage[title]{appendix}%
\usepackage{xcolor}%
\usepackage{textcomp}%
\usepackage{manyfoot}%
\usepackage{booktabs}%
\usepackage{algorithm}%
\usepackage{algorithmicx}%
\usepackage{algpseudocode}%
\usepackage{listings}%

\usepackage{xspace}
\usepackage{dsfont}
\usepackage{float}

\newcommand{\AdamMCMC}{\texttt{AdamMCMC}\xspace}
\newcommand{\Adam}{\texttt{Adam}\xspace}

\title{Calibrating Bayesian Generative Machine Learning for Bayesiamplification}

\author[a]{S.~Bieringer}
\author[b]{S.~Diefenbacher}
\author[a]{G.~Kasieczka}
\author[c]{M.~Trabs}

\affiliation[a]{Institut f\"ur Experimentalphysik, Universit\"at Hamburg, Luruper Chaussee 149, 22761 Hamburg, Germany}
\affiliation[b]{Physics Division, Lawrence Berkeley National Laboratory, Berkeley, CA 94720, USA}
\affiliation[c]{Department of Mathematics, Karlsruhe Institute of Technology, Englerstr. 2, 76131 Karlsruhe, Germany}

\emailAdd{sebastian.guido.bieringer@uni-hamburg.de}

\abstract{
Recently, combinations of generative and Bayesian deep learning have been introduced in particle physics for both fast detector simulation and inference tasks.
These neural networks aim to quantify the uncertainty on the generated distribution originating from limited training statistics.
The interpretation of a distribution-wide uncertainty however remains ill-defined.
We show a clear scheme for quantifying the calibration of Bayesian generative machine learning models.
For a Continuous Normalizing Flow applied to a low-dimensional toy example, we evaluate the calibration of Bayesian uncertainties from either a mean-field Gaussian weight posterior, or Monte Carlo sampling network weights, to gauge their behaviour on unsteady distribution edges. 
Well calibrated uncertainties can then be used to roughly estimate the number of uncorrelated truth samples that are equivalent to the generated sample and clearly indicate data amplification for smooth features of the distribution.
}


\begin{document}
\maketitle
\flushbottom

\section{Introduction}
\label{sec:intro}

The upcoming high-luminosity runs of the LHC will push the quantitative frontier of data taking to over $25$-times its current rates. 
To ensure precision gains from such high statistics, this increase in experimental data needs to be met by an equal amount of simulation.
The required computational power is predicted to outgrow the increase in budget in the coming years~\cite{HEPSoftwareFoundation:2017ggl, Boehnlein:2022}.
One solution to this predicament is the augmentation of the expensive, Monte Carlo-based, simulation chain with generative machine learning.
A special focus is often put on the costly detector simulation~\cite{Butter22_MLgeneration, Hashemi:2023rgo_generative}.

This approach is only viable under the assumption that the generated data is not statistically limited to the size of the simulated training data.
Previous studies have shown, for both toy data~\cite{Butter2020_GANplify} and calorimeter images~\cite{Bieringer2022_Calomplify}, that samples generated with generative neural networks can surpass the training statistics due to powerful interpolation abilities of the network in data space.
These studies rely on comparing a distance measure between histograms of generated data and true hold-out data to the distance between smaller, statistically limited sets of Monte Carlo data and the hold-out set. 
The phenomenon of a generative model surpassing the precision of its training set is also known as amplification.
While interesting in theory and crucial for the pursuit of the amplification approach, these studies can not be performed in experimental applications as they rely on large validation sets multiple orders of magnitude bigger than the training data.

Recently, generative architectures employing Bayesian network weight posteriors have been applied to event generation~\cite{CFM_butter2023jet} allowing the generation of sets of data with a corresponding uncertainty on the generated data distribution.
In the limit of large generated sets, this uncertainty is entirely based in the statistical limitations of the training data.
For well calibrated uncertainty predictions, this raises the question whether an estimate of statistical power of the generated data can be formed from the uncertainty prediction itself.
In this paper,
\begin{itemize}
    \item we introduce a technique for quantifying the calibration of Bayesian uncertainties on generative neural networks based on the mean coverage of the prediction.
    \item We then develop an estimate of the number of simulated truth events matching the generated set in statistical power and validate this estimate.
\end{itemize}
For applications where the uncertainty calibration can be ensured, for example by evaluating on a validation region, this approach gives an inherent quantification of the significance of a generated set.

In Bayesian neural networks (BNNs) and beyond, calibrating uncertainty quantification is crucial for correct application of the prediction results~\cite{Chen2022_UQ_review}.
While we prefer the uncertainties to align perfectly with the prediction error, overconfident predictions will lead to inflated significance values and false discoveries.
Underconfident predictions on the other hand will obscure findings, but not lead to false results and can thus be tolerated to small extend.

Bayesian generative machine learning is inherently different from other BNNs in particle physics applications such as regression~\cite{Kronheim2020_BNN_SUSY} or classification~\cite{Bollweg:2019skg, Araz2021_combine_conquer}.
Notably, in generative modeling, a low density region of data cannot be understood as low training statistics, but rather as a feature of the data that has to reproduced by the network.
The uncertainty estimate thus behaves similarly to a low-dimensional, parameterized fit~\cite{Bellagente2021_BINN} introducing high error estimates at steep features of the data distribution or whenever the function class induced by the network architecture is not sufficient to reproduce the data.
In a subsequent study of the quality of event generators~\cite{Das2023_limitations_tilman}, the authors also connect low uncertainty to good performance of the posterior mean in terms of a classifier test, but find that the weight distribution of a classifier is more sensitive to diverse failure modes than the Bayesian uncertainty.

In Section~\ref{sec:bnn}, we will explain the basic concepts of BNNs, while the connection to generative machine learning will be made in Section~\ref{sec:model}.
We introduce the toy data, as well as the employed binning in Section~\ref{sec:data} and use them to evaluate the calibration of two different classes of BNNs in Section~\ref{sec:cal}.
The idea of employing the Bayesian uncertainties for amplification is developed and deployed in Section~\ref{sec:amp}, before we conclude in Section~\ref{sec:concl}.

\section{Bayesian Neural Networks}
\label{sec:bnn}

In contrast to traditional, frequentist deep neural networks, in a Bayesian phrasing of deep learning, a distribution on the network weights is applied.
This distribution encodes the belief in the occurrence of the weight configuration $\theta$.
%rather than using an ensemble of point estimates.
This, so called \textit{posterior} distribution
\begin{equation}\label{eq:bayes_formula}
    \pi(\theta|\mathcal{D}) = \frac{\pi(\mathcal{D}|\theta)\, \pi(\theta)}{\pi(\mathcal{D})}
\end{equation}
is formed from our \textit{prior} beliefs $\pi(\theta)$ and the \textit{likelihood} $\pi(\mathcal{D}|\theta)$ of the data $\mathcal D$ under the model.
While the likelihood gives the probability of the data given its modelling through the network and thus encodes the data inherent distribution (aleatoric uncertainty), the posterior distribution provides the uncertainty due to a lack of data (epistemic uncertainty)~\cite{bayesianNN_handbook}.

Multiple methods of accessing the posterior distribution exist. 
For a broad overview over the existing techniques, we refer the readers to~\cite{Chen2022_UQ_review, bayesianNN_handbook, bayesianNN_survey, bayesianNN_survey2}.
They can mostly be classified as either approximating or sampling the posterior.

One popular option is approximating the posterior as an uncorrelated Gaussian distribution by learning a mean and a standard deviation per network weight. 
These parameters of the approximation are then inferred with (stochastic) variational inference.
This technique is also referred to as `Bayes-by-Backprop'~\cite{bbb_blundell2015weight} or within High-Energy Physics often understood as `Bayesian Neural Networks'.
We will refer to it as `Variational Inference Bayes'(VIB).

For sampling the posterior, Markov Chain Monte Carlo (MCMC) methods are employed, with full Hamiltonian Monte Carlo (HMC) often considered the gold-standard~\cite{izmailov2021bayesian_HMC}.
To adapt this class of methods to the large datasets and high dimensional parameter spaces of deep learning stochastic and gradient-based chains have been developed. 
Most notably among them are stochastic gradient HMC~\cite{chen2014sgHMC} and its variations.
Due to its easy application to different machine learning tasks and great performance on previous generative applications~\cite{Bieringer2024surrogates}, we use \AdamMCMC~\cite{bieringer2023adammcmc} as one instance of MCMC-based Bayesian inference of network weights.

With access to the posterior distribution of a neural network $f_\theta(x) = y$, we can generate the network prediction as the posterior mean prediction and its uncertainty prediction as 
\begin{equation}\label{eq:bayes_mean_std}
    \hat{y} = 
    %\int_{\theta\sim \pi (\theta|\mathcal{D})} 
    \int \mathrm{d} \theta \, \pi(\theta|\mathcal{D}) \,
    f_\theta(x) \quad \text{ and } \quad \sigma_{\hat{y}}^2 = 
    %\int_{\theta\sim \pi(\theta|\mathcal{D})} 
    \int \mathrm{d} \theta \, \pi(\theta|\mathcal{D}) \,
    \left[f_\theta(x) - \hat{y}\right]^2.
\end{equation}
Here, the integration is approximated as a summation over an ensemble of network weights obtained from the posterior directly via sampling or from its approximation.

For generative machine learning, a per-sample uncertainty cannot be evaluated due to the unsupervised setup of the problem. 
We thus generate sets of data with every network weight instance in the ensemble, calculate histograms for each set and report the mean and standard deviation per bin over all sets.
This allows us to compare against the expected truth values in each histogram bin.

\section{Bayesian Continuous Normalizing Flows}
\label{sec:model}

Generative models of various flavours have been applied for fast simulation of detector effects~\cite{Butter22_MLgeneration, Hashemi:2023rgo_generative}. 
Meanwhile, Normalizing Flows, both block-based~\cite{flows_rezende2015variational} and continuous~\cite{CNF_chen2018neural}, can be connected to Bayesian machine learning straight-forwardly, as the $\log$-likelihood of the model is accessible.
Due to the recent success of diffusion-style models in detector emulation~\cite{Mikuni2023_diffusion, Mikuni2023_diffusion_calosscore2, Leigh2023_diffusion, Buhmann2023_diffusion_caloclouds2, Buhmann2023_diffusion_epic, Kobylianskii2024_diffusion_CaloGraph} and their high data efficiency, we combine both and concentrate on Continuous Normalizing Flows (CNF) in this study.

Let $x\in \mathds{R}^d$ be a point in the data set $\mathcal D$.
Following~\cite{CFM_lipman2023flow} , we first introduce the \textit{flow} mapping $\phi_t: [0,1] \times \mathds{R}^d \rightarrow \mathds{R}^d$ parameterized by a time parameter $t\in [0,1]$.
In analogy to the application of multiple blocks in a coupling-block flow~\cite{flows_rezende2015variational}, the change of the flow mapping between target and latent space is determined by an ordinary differential equation (ODE)
\begin{equation}\label{eq:CNF_ODE}
    \frac{\mathrm{d}}{\mathrm{d} t} \phi_t(x) = v_t(\phi_t(x)),\quad \phi_0(x)=x,
\end{equation}
through a time dependent \textit{vector-field} $v_t\colon [0,1] \times \mathds{R}^d \rightarrow \mathds{R}^d$.
For Diffusion Models this differential equation is promoted to a stochastic differential equation through the addition of time-dependent noise.
For both cases, the vector-field is approximated using a deep neural network
\begin{equation*}
    \tilde v_t(\cdot,\theta)\approx v_t. 
\end{equation*}
By convention, the flow is constructed to model latent data from a standard Gaussian at $t=0$ and detector/toy data at $t=1$.
This defines the the boundaries of the probability path induced by the flow mapping
\begin{equation}\label{eq:COV}
    p_t(x) = p_0\left( \phi_t^{-1}(x)\right) \det \left( \frac{\partial \phi_t^{-1}(x)}{\partial x}\right).
\end{equation}
To circumvent solving the ODE to calculate the likelihood of the input data during training, we employ Conditional Flow Matching (CFM)~\cite{CFM_lipman2023flow}.
Instead of the arduous ODE solving, the CFM loss objective matches the neural network predictions $\tilde{v}_t(x; \theta)$ to an analytical solution $u_t$, by minimizing their respective mean-squared distance
\begin{equation}\label{eq:CFM}
    \mathcal{L}_{\mathrm{CFM}}(\theta)=\mathds{E}_{t, q(x_1), p_t(x|x_1)}\left\|u_t(x|x_1)- \tilde{v}_t(x; \theta))\right\|^2.
\end{equation}
The expectation value is calculated by sampling $t \sim \mathcal{U}(0,1)$, $x_1 \sim q$ and $x \sim p_t(\cdot|x_1)$, with $q$ the probability distribution of the detector/toy data.
An efficient and powerful choice of $u_t$ is the optimal transport path~\cite{CFM_lipman2023flow}. 
By applying a Gaussian conditional probability path the CFM loss objective reduces to 
\begin{equation}\label{eq:CFM_OT}
    \mathcal{L}_{\mathrm{CFM}}(\theta)
    = \mathds{E}_{t, q(x_1), p(x_0)}
    \Big\| \left( x_1-\left(1-\sigma_\mathrm{min}\right)x_0\right)  - \tilde{v}_t(\sigma_t x_0+\mu_t; \theta))\Big\|^2.
\end{equation}
Here, we use the conventions $\mu_t = tx_1$ and $\sigma_t = 1-(1-\sigma_\mathrm{min})t$, as well as the Gaussian latent distribution $p(x_0) = \mathcal{N}(0,1)$ and a small parameter $\sigma_\mathrm{min}$, that mimics the noise level of the training data.

\subsection{Variational Inference Bayes}\label{ssec:VIB}

The parameters of an approximation $\tilde{\pi}(\theta)$ of the posterior distribution $\pi(\theta|\mathcal{D})$ can be inferred, by minimizing their Kullback-Leibler (KL) divergence using stochastic gradient descent methods~\cite{bbb_blundell2015weight}. 
As the posterior is not analytically accessible, Bayes' theorem~\eqref{eq:bayes_formula} is employed to rewrite the KL divergence in terms of the $\log$-likelihood and the distance to the prior
\begin{equation}\label{eq:bbb}  
    \mathcal{L}_{\mathrm{VIB}}
    =D_\mathrm{KL}\left[\, \tilde{\pi}(\theta), \pi\left(\theta  | \mathcal{D}\right)\right] 
    =-\int \mathrm{d} \theta \, \tilde{\pi}(\theta) \, \log \pi\left(\mathcal{D} | \theta\right) 
    + D_\mathrm{KL}\left[\, \tilde{\pi}(\theta), \pi(\theta)\right]+\text{ constant}.
\end{equation}
The $\log$-likelihood of the data under the CNF can be directly employed here.
However, calculating the $\log$-likelihood of a CNF is costly as the ODE~\eqref{eq:CNF_ODE} needs to be solved for every point in the training data.
The authors of~\cite{CFM_butter2023jet} thus propose, to substitute the $\log$-likelihood with the CFM loss~\eqref{eq:CFM_OT} and attribute for the difference by a tunable factor $k$
\begin{equation}\label{eq:bbb_CFM}
    \mathcal{L}_{\mathrm{VIB-CFM}} = \mathds{E}_{\tilde{\pi}(\theta)}\mathcal{L}_{\mathrm{CFM}} 
    + k D_\mathrm{KL} \left[\, \tilde{\pi}(\theta), \pi(\theta)\right].
\end{equation}
Similar to changing the width of the prior $\pi(\theta)$, varying $k$ adjusts the balance of the CFM-loss to the prior and thus both the bias and variance of the predicted distributions.
Trainings at low values of $k$ produce better fits at smaller uncertainties, while higher values impact the fit performance by imposing higher smoothness at the trade-off of higher estimated uncertainties.
In our experience, promoting a CFM model to a BNN this way increases the training time considerably, due to the low impact and thus slow convergence of the KL-loss term.
Possible ways to mitigate this include initiating the prior distribution and the variational parameters from the a pretrained deterministic neural network~\cite{krishnan2020specifying_moped}.

\subsection{Markov Chain Monte Carlo}\label{ssec:adammcmc}

A competing approach to variational inference-based Bayesian deep learning is MCMC sampling. 
Our approach to MCMC sampling for neural networks, \AdamMCMC~\cite{bieringer2023adammcmc},  uses the independence of the sampled invariant distribution to the starting point to initiate the sampling from CFM-trained model parameters $\theta_0$.
This drastically reduces the optimization time over the joint optimization of Section~\ref{ssec:VIB}, and makes employing the costly $\log$-likelihood for the consequent uncertainty quantification feasible.

For every step of the chain, the ODE~\eqref{eq:CNF_ODE} is solved to determine the negative $\log$-likelihood $\mathcal{L}_\mathrm{NLL}$ of the data to construct a chain drawn from a proposal distribution around an $\Adam$~\cite{adam_Kingma2014AdamAM} step
\begin{equation}
    \tilde\theta_{i+1} = \Adam(\theta_i,  \mathcal{L}_{\mathrm{NLL}}(\theta_i)).
\end{equation}
In combination with a proposal distribution that is elongated in the direction of the step
\begin{equation}
    \tau_i \sim q(\cdot|\theta_i) = \mathcal{N}(\tilde\theta_{i+1},\sigma^2\mathrm{1}+\sigma_\Delta(\tilde\theta_{i+1} - \theta_i)(\tilde\theta_{i+1} - \theta_i)^\top),
\end{equation}
this algorithm handles high dimensional sampling for neural networks very efficiently and results in a high acceptance rate in a subsequent stochastic Metropolis-Hastings (MH) correction with acceptance probability
\begin{equation}
    \alpha = \frac{\exp \left(-\lambda \mathcal{L}_{\mathrm{NLL}}(\tau_i)\right) q(\theta_i | \tau_i)}{\exp \left(-\lambda \mathcal{L}_{\mathrm{NLL}}(\theta_i)\right) q(\tau_i | \theta_i)}
\end{equation}
for a large range of noise parameter settings.
If the added noise $\sigma$ is low, the results remain close to the stochastic optimization without error estimates close to zero, but if the noise levels are high, the random walk through parameter space dominates and the algorithm does not converge to a sensible parameter values.
This behaviour is masked by diminishing acceptance probabilities for very low and very high $\sigma$~\cite{bieringer2023adammcmc}.

Both the inverse temperature parameter $\lambda$ and the noise parameter $\sigma$ tune the predicted uncertainties.
In theory small $\lambda$ and high $\sigma$ will result in high error estimates, albeit in practice the dependence on the inverse temperatures is very weak.
We thus limit ourselves to adapting the noise parameter to align the generated uncertainties.

After an initial burn-in period, which can be skipped when initializing from a pretrained model, repeatedly saving the network parameters after gaps of length $l$ ensures approximately independent parameter samples.
The set of sampled parameters 
\begin{equation}
    \mathbf{\Theta}_\mathrm{MCMC} = \{\theta^{(1)},...,\theta^{(n_\mathrm{MCMC})}\} := \{\theta_{1\cdot l},...,\theta_{n_\mathrm{MCMC}\cdot l}\}
\end{equation} 
follows the tempered posterior distribution, due to Bayes‘ theorem and the resulting proportionality
\begin{equation}
\label{eq:gibbs}
    %p_\lambda (\theta|\mathcal{D}) \propto \exp\left(-\lambda \mathcal{L}_{\mathrm{NLL}}(\theta)\right).
    \tilde{\pi}_\lambda (\theta|\mathcal{D}) \propto \exp\left(-\lambda \mathcal{L}_{\mathrm{NLL}}(\theta)\right) \pi(\theta).
\end{equation}

\section{Toy Setup}
\label{sec:data}

\begin{figure}[t]
    \centering
    \includegraphics[width=\textwidth]{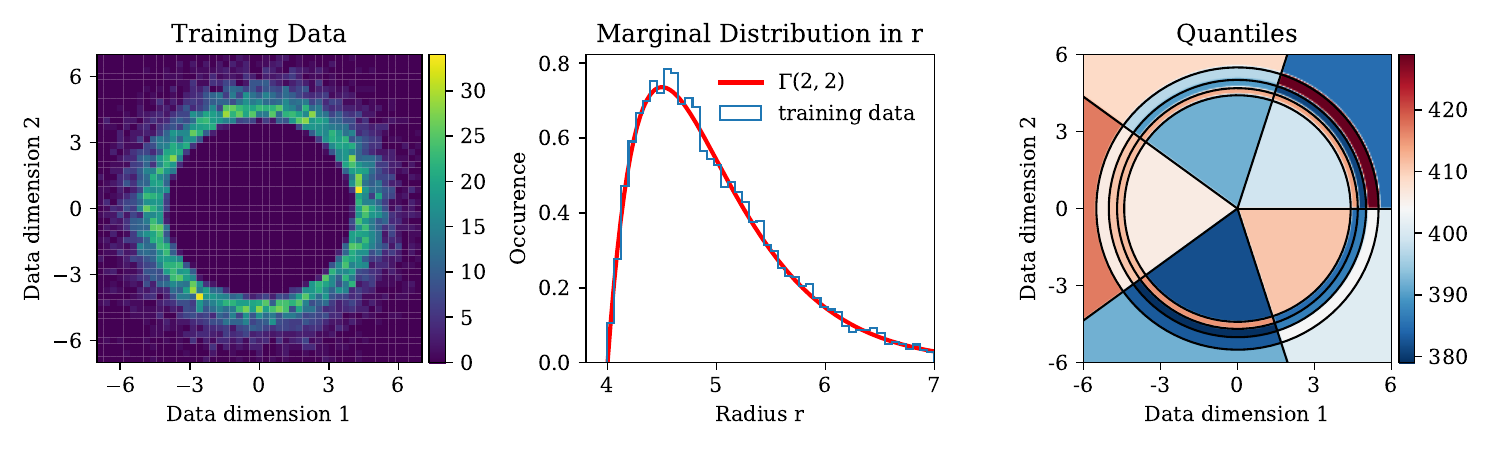}
    \caption{Left: Histogram of one training data set ($10,000$ points). 
    The data follows a ring structure with a sharp edge at $r=4$ and a long tail to higher radii.
    Mid: Marginal distribution of the training data in radial direction.
    Right: $5\times5$ quantiles generated from a data set of $10M$ points and filled with the training data. 
    The quantiles are constructed with equal probability of the truth data to fall into every quantiles.}
    \label{fig:data_and_quantiles}
\end{figure}

\subsection{Gamma Function Ring}
Similar to previous studies on data amplification~\cite{Butter2020_GANplify}, we employ the CNF on a low-dimensional ring distribution.
Generative Architectures often struggle with changes in the topology between latent space, typically Normal distributed, and data space~\cite{Winterhalder_2021_latent_space_refinement_and_obstructions}.
The ring structure reflects this ``topological worst case''.
A generalization of the results from a similar, topologically complicated, but low-dimensional toy to high-dimensional, simulated, and topologically less problematic calorimeter images was performed in \cite{Bieringer2022_Calomplify}. 
In this paper, we focus on the calibration of generative uncertainties and draw a connection to data amplification.
We thus limit the study to two dimensions for illustrative purposes and to reduce computational costs.
Nevertheless, the calibration can be executed analogously for higher dimensional distributions.
We generate samples from a ring distribution with an unsteady edge at a radius of $r=4$, by sampling in spherical coordinates from
\begin{equation*}
    \phi \sim \mathrm{uniform}(0,2\pi) \quad \text{ and } \quad r-4 \sim \Gamma(\alpha,\beta) %p(\cdot; \alpha, \beta) = \frac{(x-4)^{\alpha-1} e^{-\beta (x-4)} \beta^\alpha}{\Gamma(\alpha)}
\end{equation*}
with parameters $\alpha = \beta = 2$ for the Gamma distribution. 
Per training, we use an independent sample of $N = 10,000$ points.
Before passing the data to the CNF, we transform into Cartesian coordinates to obtain the ring shape shown in Figure~\ref{fig:data_and_quantiles}.
This construction allows us to estimate the behaviour of the uncertainties at distribution edges and simultaneously prevents divergences of the probability distribution in $(x,y)=(0,0)$.

\subsection{Hyperparameter Choices}
\label{ssec:params}

Due to the low dimensionality of the toy example, we do not need to employ complicated architectures to obtain a good approximation of the vector-field $\tilde v_t(\cdot,\theta)$. 
Based on a small grid search, a Multi-Layer Perceptron with 3 layers of 32 nodes and ELU activation is sufficient to reproduce the training data well. 
Each of the 3 layers takes the time variable $t$ as an additional input.
The neural network part of the CNF thus totals a mere $2498$ parameters.

When parameterizing the weight posterior approximation $\tilde\pi(\theta)$ as an uncorrelated Normal distribution, as is standard in VIB~\cite{bbb_blundell2015weight}, the number of parameters consequently doubles.
For VIB we train using the \Adam optimizer~\cite{adam_Kingma2014AdamAM} at a learning rate of $10^{-3}$ for up to $250$k epochs of $10$ batches of $1000$ datapoints each.
To prevent overfitting, we evaluate the model at the earliest epoch after convergence of the KL-loss term.
This point depends on the choice of $k$ and varies between $75$k for $k=50$ and $250$k for $k=1$.
We do $5$ runs each for multiple values of $k\in [1,5,10,50]$ to regulate the uncertainty quantification.
For this range of $k$ we have previously found sensible density estimation and optimization convergence trough performing a $\log$-linearly spaced scan in $k \in [10^{-4}, 10^5]$) with only one run per parameter choice.

For the \AdamMCMC sampling, we start the chain from a pretrained model.
The model is first optimized for $2500$ epochs (\Adam with learning rate of $10^{-3}$) using only the CFM-loss~\eqref{eq:CFM_OT}.
For the deterministic model, this is enough to converge.
We then run the sampling at a the same learning rate as the optimization with $\sigma_\Delta \approx 50$ and $\lambda = 1.0$.
This choice of $\sigma_\Delta$ ensures high acceptance rates, while the choice of $\lambda$ reflects sampling from the untempered posterior distribution, as per \eqref{eq:gibbs}.
We add a sample to the collection at intervals of $100$ epochs, to ensure the independence of the sampled weights.
To adjust the calibration, we scan the noise value at four points  $\sigma\in [0.01, 0.05, 0.1, 0.5]$.
This parameter span is based on a $\log$-linearly spaced scan in $\sigma \in [10^{-4}, 10]$).
Once again, we calculate $5$ chains per noise parameter setting.

\subsection{Quantiles}

As in Reference~\cite{Butter2020_GANplify}, we evaluate the generated data in histogram bins of equal probability mass.
We will refer to these bins as quantiles $Q_j$, their count as $q_j$ and the set of all quantiles as $\mathbf{Q} = \{Q_1,...,Q_{n_Q}\}$.
To construct bins with the same expected occupancy, we use spherical coordinates.
In angular direction, the space can simply be divided into linearly spaced quantiles, while in radial direction we use the quantiles of a $10$M generated truth dataset to gauge the boundaries of the quantiles.
To guaranty even population, we always choose the same number of quantiles in both dimensions.
Figure~\ref{fig:data_and_quantiles} illustrates the construction and occupancy for $5\times5$ quantiles in Cartesian coordinates.

For correlated data, quantiles can be constructed by iteratively dividing a truth set into sets of equal size~\cite{Bieringer2022_Calomplify}. 
The binning is however not relevant for the discussion of calibration and analogous arguments can be made for arbitrary histograms.
The advantage of quantiles over other binning schemes is the clear definition of the number of bins without an offset by an arbitrary amount of insignificant bins in the sparsely or unpopulated areas of the data space. 
This allows us to show the behaviour of calibration and amplification over the number of bins in Section~\ref{sec:cal} and~\ref{sec:amp}.

\section{Calibration}
\label{sec:cal}

To align the uncertainty quantification, for \AdamMCMC we generate $10$M points from the CNF for the $n_\mathrm{MCMC} = 10$ parameter samples in $\mathbf{\Theta}_\mathrm{MCMC}$.
We obtain a set of points $\mathbf{G}^{(i)}$ per parameter sample $\theta^{(i)}$, with the corresponding count 
\begin{equation*}
    g_j^{(i)} = \#\{ x' \in Q_j \mid x' \in \mathbf{G}^{(i)}\}
\end{equation*}
in quantile $Q_j$. 
Each count corresponding to a parameter sample thus constitutes one drawing of a random variable $G_j$ whose distribution is induced by the posterior.

Analogously, for VIB we draw a set $\mathbf{\Theta}_\mathrm{VIB}$ of parameters from the posterior approximation $\tilde{\pi}(\theta)$, generate $10$M samples from each and calculate the quantile counts to generate drawings of $G_j$.
As the training cost does not depend on the number of draws for VIB, we use $n_\mathrm{VIB} = 50$ samples for better accuracy.

Using the quantile values $g_j^{(i)}$, we approximate the cumulative distribution function (CDF)
\begin{equation}
    \hat{F}_{G_j,\mathbf{\Theta}}\left(g_j\right)
    \approx F_{G_j}\left(g_j\right) = P\left(G_j\leq g_j\right),
\end{equation}
from its empirical counterpart using linear interpolation.
We leave the set $\mathbf{\Theta}$ general, without a subscript, for now.
From the approximated CDF, we construct symmetric confidence intervals for a given confidence level $c$ from its inversion 
\begin{equation}
    I_{j,\mathbf{\Theta}}(c) = \left[\hat{F}_{G_j,\mathbf{\Theta}}^{\, -1}\left(0.5-\frac{c}{2}\right),  \hat{F}_{G_j,\mathbf{\Theta}}^{\, -1}\left(0.5+\frac{c}{2}\right) \right].
\end{equation}
The chosen confidence level $c$ corresponds to the expected or \textit{nominal coverage}.

To evaluate the observed coverage, we draw $5$ different training sets from the Gamma ring distribution and calculate a VIB- and \AdamMCMC-CNF ensemble each 
\begin{equation*}
    \mathbf{\Theta}_\mathrm{MCMC}^{\, s} \text{ and } \mathbf{\Theta}_\mathrm{VIB}^{\, s} \text{ for } s \in \{1,..,5\}.
\end{equation*}
For every model, we construct a confidence interval and evaluate the number of intervals containing the expected count of the truth distribution, i.e. $1/n_Q$.
The ratio of models with an interval containing the truth value over the total number of models gives the \textit{empirical coverage} per bin
\begin{equation}
    \hat{c}_j = \frac{\#\left\{ 1/n_Q \in I_{j,\mathbf{\Theta}^{\, s}} (c)\mid s \in \{1,..,5\} \right\}}{5},
\end{equation}
where we again keep the subscript on the set of parameters unspecified.
For one quantile this coverage estimate is very coarse as it can only take on one of six values.
Since we want to check the agreement of nominal and empirical coverage for multiple nominal coverage values, we report the mean empirical coverage
\begin{equation}\label{eq:mean_ec}
    \bar{c} = \left\langle \hat{c}_j \right\rangle_{j \in \{1,...,n_Q\}}
\end{equation}
over all quantiles.
The range of possible mean values is big enough to compare to a fine spacing in nominal coverage.

\begin{figure}[b]
    \centering
    \includegraphics[width=\textwidth]{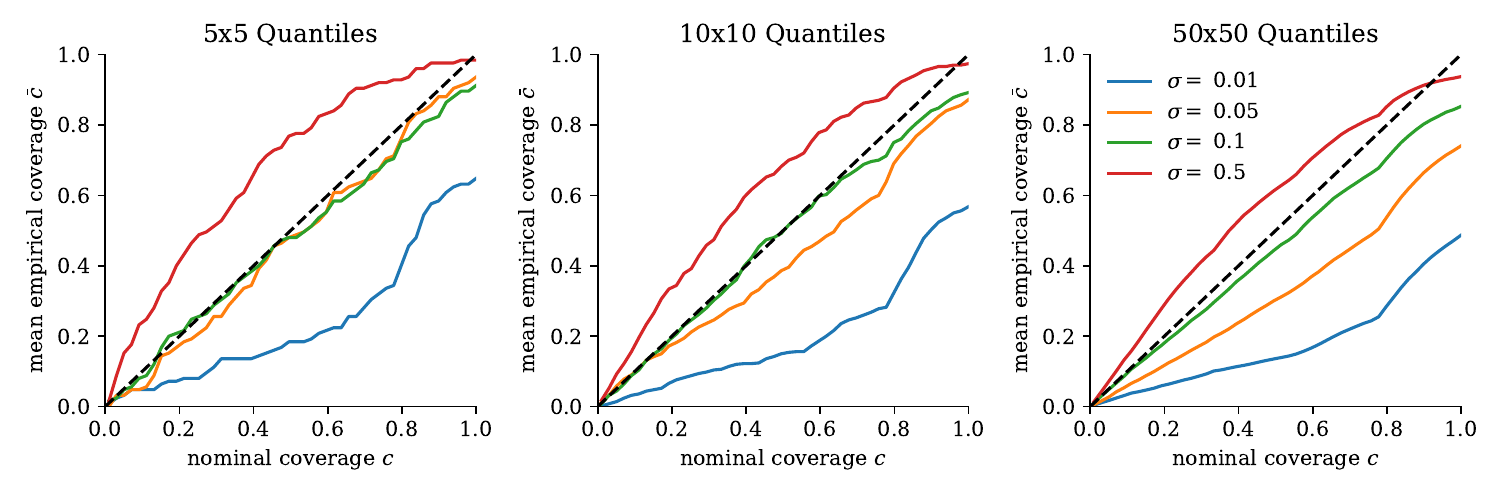}
    \caption{Mean empirical coverage for confidence intervals calculated from $10$ samples of the Bayesian weight posterior drawn with \AdamMCMC using 4 different hyperparameter settings.
    Higher $\sigma$ will generally result in larger uncertainties.
    The empirical coverage is calculated from $5$ independent runs and averaged over all quantiles.
    The panels show a clear dependence of the calibration on the number of quantiles increasing from left to right.}
    \label{fig:cal_MCMC}
\end{figure}

This also allows us to judge the agreement of nominal and empirical coverage in the full data space in a single figure. 
However, it also introduces the possibility for over- and underconfident areas to cancel each other out.
This issue will be treated in more detial in Section~\ref{ssec:nbins} and ~\ref{ssec:edge}.

Figure~\ref{fig:cal_MCMC} shows the mean empirical coverage over all quantiles for $50$ values of the nominal coverage linearly spaced between $0$ and $1$ and over three different numbers of quantiles.
For a well calibrated uncertainty estimation, the empirical estimate closely follows the nominal coverage and the resulting curve is close to the diagonal of the plot.
For Figure~\ref{fig:cal_MCMC} we can see that high noise levels in the MCMC chain lead to overestimated errors and a prediction that is underconfident on average.
Inversely, low noise levels lead to overconfident predictions.
From our chosen grid, $\sigma = 0.1$ shows the best agreement.

It further becomes apparent that the calibration depends on the number of quantiles.
For lower numbers of quantiles, the fluctuations in the generated distribution average out and both the mean prediction and error estimation are more precise, while for higher numbers of quantiles good calibration becomes challenging while limited to $10$ posterior samples.

\begin{figure}
    \centering
    \includegraphics[width=\textwidth]{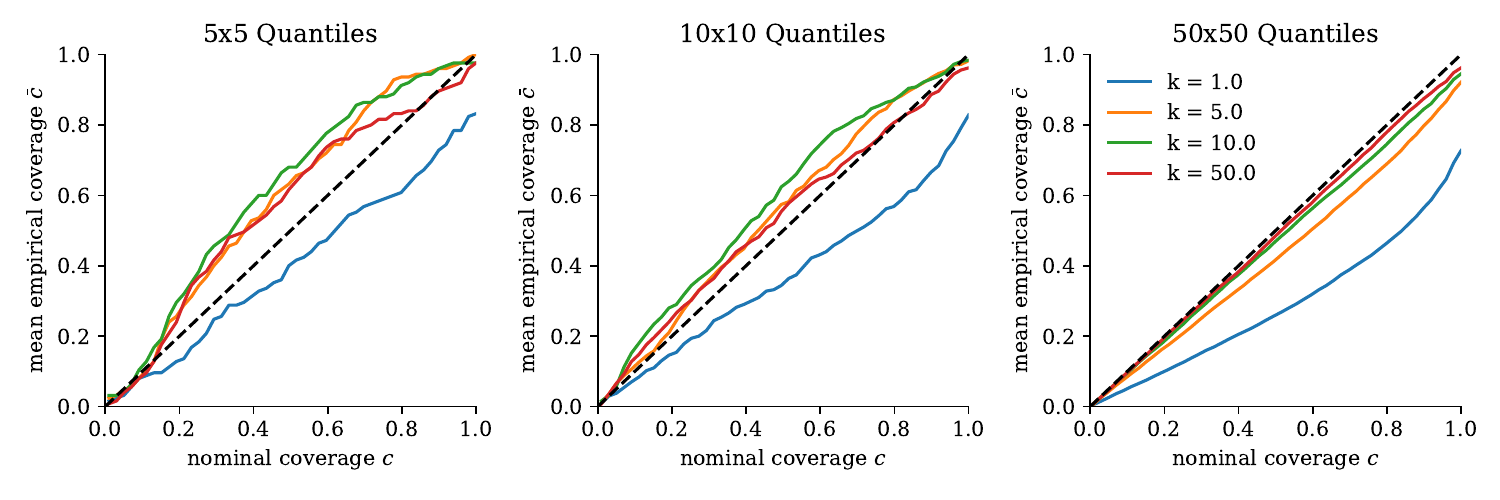}
    \caption{Mean empirical coverage for confidence intervals calculated from $50$ drawings from the VIB approximation of the Bayesian weight posterior with 4 different hyperparameter settings. 
    Larger $k$ increases the dependence of the fit on the prior.
    The empirical coverage is calculated from $5$ independent runs and averaged over all quantiles.
    The panels again show a clear dependence of the calibration on the number of quantiles increasing from left to right.}
    \label{fig:cal_VIB}
\end{figure}

For VIB in Figure~\ref{fig:cal_VIB}, where we evaluate $50$ posterior samples, calibration seems to improve for high $n_Q$.
While at lower numbers only a very small prior trade-off $k$ leads to overconfident intervals and larger values result in underconfident predictions, at higher numbers of quantiles previously underconfident predictions appear well calibrated.

\subsection{Scaling with the Number of Quantiles}
\label{ssec:nbins}

To further investigate the calibration of our Bayesian generative neural networks, we pick the seemingly best calibrated parameter settings for both methods.
For \AdamMCMC this is $\sigma = 0.1$ and for VIB $k=10$.
We generate $n_\mathrm{MCMC} = n_\mathrm{VIB} = 50$ samples from the posterior for both methods now and evaluate the scaling with the number of quantiles in more detail.

As we do not want to evaluate one calibration plot for each quantile, we reduce the diagonal calibration plots by calculating the mean (absolute) deviation between empirical and nominal coverage
\begin{equation}
    \mathrm{MD} = \left\langle \bar{c} - c \right\rangle_{c \in [0,1]}
    \quad \text{ and } \quad
    \mathrm{MAD} = \left\langle | \bar{c} - c | \right\rangle_{c \in [0,1]},
\end{equation}
where the mean empirical coverage still depends on the nominal coverage $\bar{c} = \bar{c}(c)$.
The composition of the mean on the quantiles, the absolute value, and the mean on the nominal coverage allows for under- and overestimation in individual quantiles to cancel out.

To gauge this we promote the index over all quantiles $j$ to a tuple of indices $(j_r,j_\phi)$.
We write $\hat{c}_{(j_r,j_\phi)}$ for the empirical coverage in the $j_r$-th radial and $j_\phi$-th angular bin.
By limiting the average over the empirical coverage~\eqref{eq:mean_ec} to one of the dimensions
\begin{equation}\label{eq:1d_mean_ec}
    \bar{c}_{j_r} = \left\langle \hat{c}_{(j_r,j_\phi)} \right\rangle_{j_\phi}
    %\in \{1,...,\sqrt{n_Q}\}}
    \quad \text{ and } \quad
    \bar{c}_{j_\phi} = \left\langle \hat{c}_{(j_r,j_\phi)} \right\rangle_{j_r}
    %\in \{1,...,\sqrt{n_Q}\}},
\end{equation}
we construct marginal coverage distributions in the remaining directions.
%radial dimension $\bar{c}_r$, where $r$ are intervals of the quantiles in radial direction. 
We can then again calculate the mean absolute deviation to the nominal coverage and suspend the average in the remaining direction until the very end
\begin{equation*}
    \mathrm{MAD}_r = \left\langle \left\langle | \bar{c}_{j_r} - c | \right\rangle_{c \in [0,1]} \right\rangle_{j_r}
    %\in \{1,...,\sqrt{n_Q}\}}
    \quad \text{ and } \quad
    \mathrm{MAD}_\phi = \left\langle \left\langle | \bar{c}_{j_\phi} - c | \right\rangle_{c \in [0,1]} \right\rangle_{j_\phi}
    %\in \{1,...,\sqrt{n_Q}\}}.
\end{equation*}
By switching the order, only quantile counts in direction of the first mean~\eqref{eq:1d_mean_ec} can even out.
When starting out with the average in the angular dimension, we end up with an estimate where the fluctuations in the radial direction are preserved in the absolute mean and vice versa.

\begin{figure}
    \centering
    \includegraphics[width=\textwidth]{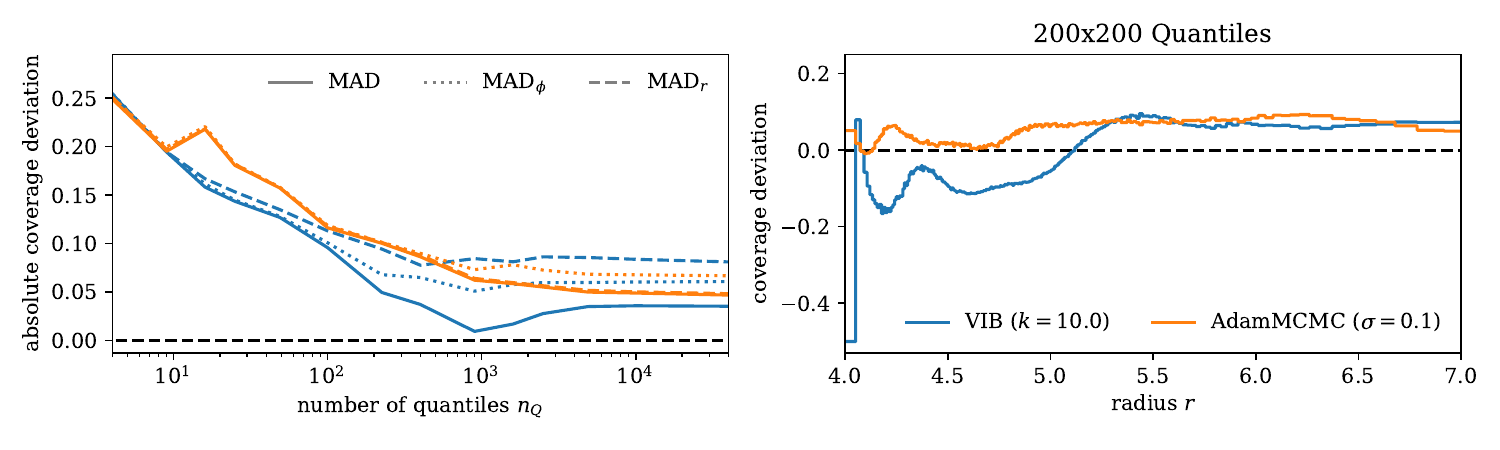}
    \caption{Left: Mean absolute deviation between the nominal and the empirical coverage ($5$ runs) for $50$ posterior samples from both, VIB at $k=10$ and \AdamMCMC at $\sigma=0.1$.
    The panel shows a strong dependence on the number of quantiles.
    From evaluating the calibration plots for all numbers of quantiles individually, we know both methods are undercertain at low numbers. 
    With increasing $n_Q$, the calibration mean exhibits a strong dependence on the order of the absolute and average operations.
    When only the over- and underdensities along radial or angular dimension can cancel out, the mean calibration of the VIB-CNFs drastically decreases.
    Right: Difference between nominal and empirical coverage (mean over angular direction) for the radial direction of $200\time200$ quantiles.
    Values below $0$ indicate overcertainty, while values above indicate undercertain predictions.
    While the predictions of the \AdamMCMC-CNFs are well calibrated to slightly undercertain along the radius, the prediction of the VIB-CNFs starts out overcertain due to the oversmoothing caused by the strong prior dependence.
    When averaging over all radii, the VIB-CNF predictions cancel exactly and the BNN seems well calibrated.}
    \label{fig:scaling}
\end{figure}

The left panel of Figure~\ref{fig:scaling} shows the dependence of the three different coverage deviation averages on the number of quantiles.
To keep the effect of statistic fluctuations per bin to a minimum, we generate sets of $1000\cdot n_Q$ artificial points with the CNFs and evaluate between $2\times 2$ and $200\times 200$ quantiles.

We find a clear dependence of the coverage means on the number of quantiles.
At low numbers, the mean prediction averages over large areas of the data space increasing the quality of the mean prediction.
The uncertainty estimation is thus underconfident for both methods and the mean absolute deviations are high and do not depend on the order of  averaging.
For low numbers of quantiles, calibration is much better,i.e. the absolute deviation is closer to $0$. 
While for \AdamMCMC we cannot see big changes depending on the averaging order. 
This indicates a calibration independent of the dimension. 
At the same time, we find large discrepancies for VIB.
This variation can be understood from the marginal calibrations.

\subsection{Calibration at Sharp Features in Radial Direction}
\label{ssec:edge}

The right panel of Figure~\ref{fig:scaling} displays the the marginal empirical coverage in radial direction $\bar{c}_r$ for $200\times200$ quantiles and both BNN methods.
We can see a distinct difference in the uncertainty quantification.

While the VIB prediction seems very well calibrated in total, in the radial direction, the VIB underestimates its bias for the steeply rising part of the data distribution between $r\in [4.0,5.0]$.
For the same interval, the MCMC prediction is well calibrated and less underconfident than for higher radii.
For $r>5$ both models slightly overestimate the uncertainty and show very similar calibration.

In terms of absolute uncertainty, both methods actually predict very similar results.
However, the mean prediction of the VIB-CNF is strongly biased by the prior KL-loss term, resulting in large underpopulation of the generated density due to oversmoothing for $r<4.5$ and a corresponding overpopulation in $r\in [4.5,5.0]$.
We have tested the predictions for $k=50$ and the behaviour is magnified at higher values of $k$.
For lower values ($k=5$), the oversmoothing is reduced to the area below $r=4.3$ at the cost of an overestimated tail.
The \AdamMCMC-CNF shows signs of oversmoothing as well, but only very close to the start of the radial distribution.

\section{Bayesiamplification}
\label{sec:amp}

Based on the previous discussion of both the total and marginal calibration, we can confidently say that our \AdamMCMC-CNF is well calibrated, albeit slightly underconfident for some areas of data space and small numbers of bins.
It is, however, important to note that truth information was needed  to evaluate the calibration of the BNN. 
In a practical application, this would require either a validation region or a large hold-out set, the latter of which would partially defeat the purpose of data amplification in fast detector simulation.
However, for applications with validations regions, such as generative anomaly detection~\cite{Hallin2021_cathode, Golling2023_cathode_compare}, precision improvements through data amplification can be realized.

With a well calibrated BNN, we can try and develop a measure of the statistical power of the generated set from the uncertainties.
We do so by relating the uncertainty to the statistics of an uncorrelated set of points $\mathbf{T}$ from the truth distribution.
For $n_\mathrm{bins}$ arbitrary bins, we expect the count in the $j$-th bin to be approximately Poisson distributed with mean and variance $t_j$.
For the same bin, the set of $n_\mathrm{MCMC}=50$  \AdamMCMC-CNF posterior samples gives a mean prediction
\begin{equation*}
    \bar{g}_j = \left\langle g_j^{(i)}\right\rangle_{i \in \left\{1,...,n_\mathrm{MCMC}\right\}}
    \text{ and  variance } \sigma_{\bar{g}_j}^2 = \left\langle \left( g_j^{(i)} - \bar{g}_j  \right)^2 \right\rangle_{i \in \left\{1,...,n_\mathrm{MCMC}\right\}}.
\end{equation*}
We will now use the posterior mean and variance to construct an estimator $\hat{t}_j$ of the Poisson equivalent to the per-bin predictions.
Using only the mean $\hat{t}_j := \bar{g}_j$, the equivalent will simply be the generated statistics.
Thereby, we would disregard the correlations in the generated data through limited training data completely.

By instead equating the variance of the BNN to that of the equivalent uncorrelated set $\hat{t}_j~:= ~\sigma_{\bar{g}_j}^2$, 
we would introduce an unwanted dependence on the uncertainty prediction. 
Overestimated uncertainties would lead to an overestimation of the statistical power.

As we do not want to overestimate the generative performance, we aim to have undercertain predictions to lead to an underestimation of the uncorrelated equivalent. 
Such a behaviour can be constructed using the coefficient of variation
\begin{equation}
    \frac{1}{\sqrt{\hat{t}_j}} := \frac{\sigma_{\bar{g}_j}}{\bar{g}_j}
    \quad \Longleftrightarrow \quad
    \hat{t}_j = \frac{\bar{g}_j^2}{\sigma_{\bar{g}_j}^2}.
\end{equation}
The equivalent uncorrelated statistics now decreases for overestimated $\sigma_{\bar{g}_j}$. 
Both the predictions from the absolute and from the relative error give the similar estimates for well calibrated errors in our tests.

\begin{figure}[t]
    \centering
    \includegraphics[width=\textwidth]{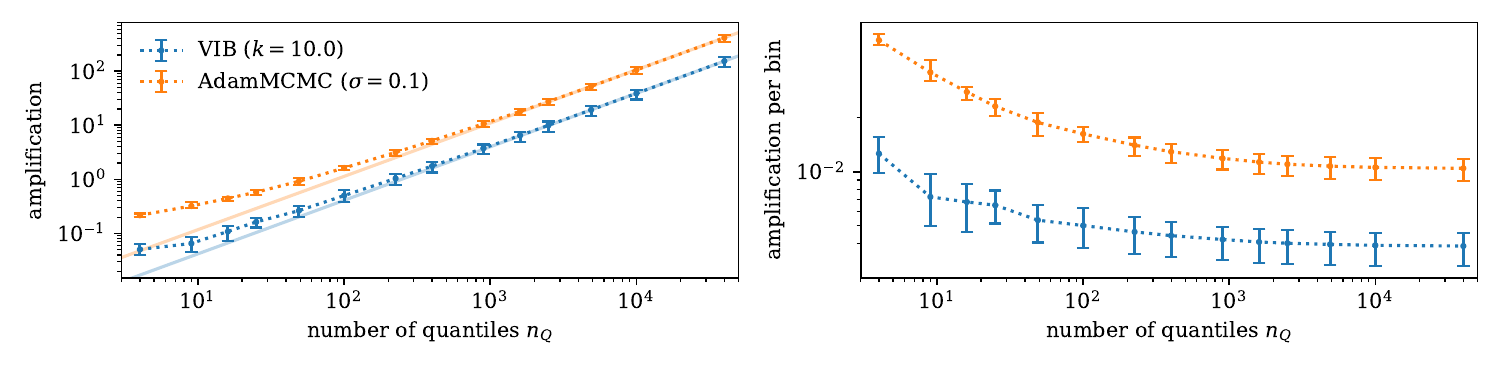}
    \caption{Left: Amplification estimate generated by equating the error prediction per bin for both BNNs to the Poisson error of an independent data set.
    Higher uncertainty results in lower amplification.
    Errorbars are calculated from the ensemble of $5$ runs done per BNN method.
    Again, we use $50$ samples of the weight posterior (approximation) for both methods.
    The faint solid lines show the result of exponential linear fit (least squares) to the last $8$ points.
    Right: The mean estimate over all quantiles converges to a constant value at high numbers of quantiles resulting in linear scaling of the amplification with the number of quantiles.
    }
    \label{fig:amp}
\end{figure}

We calculate the equivalent truth set size for both the VIB-CNF and \AdamMCMC-CNF and the quantiles from Section~\ref{sec:cal}.
In Figure~\ref{fig:amp}, we report the \textit{amplification} as the ratio of the sum over all bin estimates and the training statistics \[\sum_{j=1}^{n_\mathrm{bins}} \hat{t}_j/N\] in the left panel, as well as the mean estimate over all bins on the right.

Since the amplification contains the sum over all quantiles of our setup and $\hat{t}_j$ depends on the fluctuations of the individual predictions $g_j^{(i)}$ around the posterior mean prediction only, we expect it to scale linearly in the number of bins.
This seems in good agreement with the figure.
For large numbers of quantiles, where the BNNs are best calibrated, the average amplification per bin converges to a constant value.
Fitting a exponential linear function $\exp(a+b\cdot \log(x)) = a'\cdot x^b$ to these last $8$ points of Figure~\ref{fig:amp} using least squares, we indeed find no significant deviations from $b=1$.
We estimate $a' = (4.3 \pm 2.9)\cdot10^{-3}$ and $b=0.99 \pm 0.06$ for the VIB-CNF and $a' = 0.012 \pm 0.004$ and $b=0.99 \pm 0.04$ for the \AdamMCMC-CNF.
At lower numbers, the deviations of the model output for different parameters in the Bayesian set integrate over large intervals of the data space leading to smaller error estimates and increased amplification per bin.

This behaviour is consistent with the previous studies~\cite{Butter2020_GANplify, Bieringer2022_Calomplify} and the observation that one can not improve the estimation of low moments of the distribution, like the distribution mean, by oversampling with a generative neural network.
From Figure~\ref{fig:amp}, we can also estimate the minimum amount of bins to leverage the amplification.   
For the MCMC sample, evaluating at $100$ bins is expected to yield an improved density estimation over using only the training set.
This number could decrease for a less underconfident model.
For highly granular binning, we find amplification estimates of more than a factor $100$.

For smaller training statistics, we expect a higher initial amplification at low numbers of bins, while the corresponding larger uncertainty estimate will result in a flatter slope.
The number of quantiles where an amplification larger than $1$ first occurs will be smaller in such a case.
Higher training statistics on the other hand will lead to a steeper slope and a later trade-off point.
The results of~\cite{Butter2020_GANplify} imply, that amplification effects are stronger in larger data spaces, due to the reduced density of the training data.

Similar calculations can be done for arbitrary binnings to justify the use of generative machine learning in a specific analysis.
The evaluation of the Bayesian uncertainty prediction however requires the calculation of multiple sets of fast-simulation data points.
This reduces the speed benefits of applying generative machine learning over more classical tools like MCMC simulation or inference.

\subsection{Checking Amplification with Jensen-Shannon Divergence}
\label{ssec:calomplify}

To test how well the sum over all bin estimates
\[\hat{N} = \sum_{j=1}^{n_\mathrm{bins}} \hat{t}_j\]
actually gauges the size of an equivalent independent data set, we calculate the Jensen-Shannon (JS) divergence
\begin{equation}
            \bar{D}_\mathrm{JS} (p,q) = \frac{1}{2} \sum_{j=1}^{n_\mathrm{bins}} \left( p_j \log \frac{p_j}{\frac{1}{2}(p_j + q_j)} + q_j \log \frac{q_j}{\frac{1}{2}(p_j + q_j)}\right)
\label{eq:bjsd}
\end{equation}
between the histogram estimation of the density in our quantiles and the known data distribution.
The JS divergence is bounded by $0$ and $\log{2}$, with smaller values indicating similarity between the compared distributions.

\begin{figure}
    \centering
    \includegraphics[width=\textwidth]{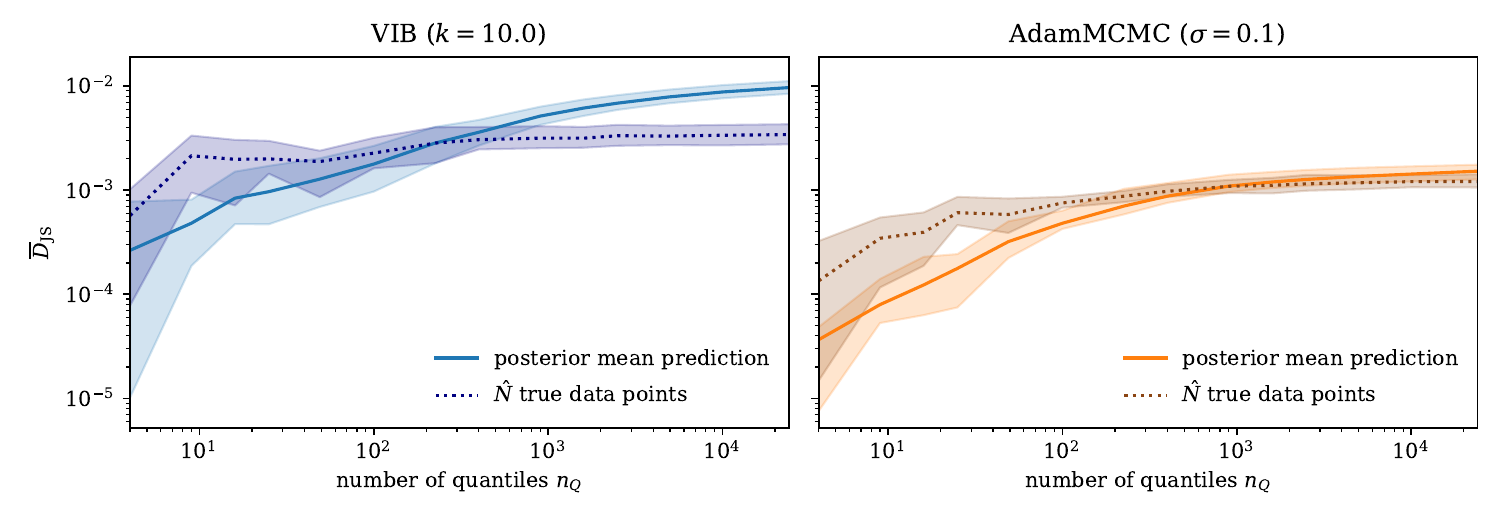}
    \caption{Jensen-Shannon divergence between the mean prediction by the Bayesian CNFs and the true data distribution (solid line) and between a uncorrelated data set of the size predicted by the BNN error estimate and the true data (dashed line).
    Both divergences align only for well calibrated uncertainties (\AdamMCMC at high numbers of quantiles) and indicate over- and underestimated errors otherwise.
    Uncertainties are reported from the ensemble of $5$ independent runs done for estimation of the empirical coverage.
    }
    \label{fig:closure}
\end{figure}

In our toy setup, the bins are constructed as quantiles. 
We evaluate the JS divergence for $p_j = \frac{\bar{g}_j}{1000\cdot n_Q}$, the mean prediction of the BNN relative to total number generated, and $q_j = 1/n_Q$ the probability per quantile when sampling from the data distribution.
In Figure~\ref{fig:closure}, we compare it to the JS divergence for $p_j = t_j/\hat{N}$, the relative population of the quantiles for a set of $\hat{N}$ points drawn from the truth distribution, and the true quantile count $q_j = 1/n_Q$ for a large range of $n_Q$.

Where the BNN is well calibrated, i.e. for \AdamMCMC and $n_Q~>~10^3$, the quality of the mean prediction lines up with the results of the uncorrelated set drawn to the size of the BNN errors.
The Bayesian coefficient of variation correctly predicts the equivalent uncorrelated statistics.
At lower numbers of quantiles, the error is overestimated.
Consequently, the statistical equivalent is underestimated.
This can also be observed for the VIB-CNF.
However, for large number of quantiles where the uncertainty at low radii is underestimated, see Section~\ref{ssec:edge}, the performance of the mean prediction is worse than anticipated by the BNN.
Good calibration on the full data space therefore is important for a reliable prediction of $\hat{N}$.

\section{Conclusion}
\label{sec:concl}

In the previous chapters, we present a novel evaluation of the uncertainty provided by a Bayesian generative neural network in a histogram.
To this end, we propose constructing confidence intervals per histogram bin and compare the nominal coverage of the constructed interval to the empirical coverage obtained from a small ensemble of BNNs.

We observe a strong dependence of the calibration on the parameters of both a VIB-CNF and an MCMC-sampled CNF.
Furthermore, we find a strong tendency to oversmooth with strong priors leading to underestimation of the data density and corresponding error at the non-differentiable inner edge of our toy distribution.
While present in both approaches, this behavior was predominantly displayed by the VIB-CNFs.

We further use the calibrated errors to estimate the statistical power of the generated data in terms of the size of an equivalent independently sampled data set.
This estimate correctly quantifies the performance of the BNNs mean prediction when the errors are well calibrated and assigns a concrete number to the data amplification in dependence of the employed binning.
For a correct amplification estimate, it is crucial that the errors are well calibrated in the full data space.

Similar calibration checks can be applied wherever a generative neural network is used for inference or generation with a sufficient validation set or for interpolation into hold-out regions of the data.

% \appendix
% \section{Some title}
% Please always give a title also for appendices.

\acknowledgments

SB is supported by the Helmholtz Information and Data Science Schools via DASHH (Data Science in Hamburg - HELMHOLTZ Graduate School for the Structure of Matter) with the grant HIDSS-0002.
SB and GK acknowledge support by the Deutsche Forschungsgemeinschaft (DFG) under Germany’s Excellence Strategy – EXC 2121  Quantum Universe – 390833306 and via the KISS consortium (05D23GU4) funded by the German Federal Ministry of Education and Research BMBF in the ErUM-Data action plan.
This research was supported in part through the Maxwell computational resources operated at Deutsches Elektronen-Synchrotron DESY, Hamburg, Germany. 
SD is supported by the U.S. Department of Energy (DOE), Office of Science under contract DE-AC02-05CH11231.

\paragraph{Data and Code} \url{https://github.com/sbieringer/Bayesiamplify} provides the code for simulating the toy example and conducting this analysis.

% Bibliography

\bibliographystyle{unsrt}
\bibliography{biblio.bib}

\end{document}